\documentclass[final]{cvpr}
  \usepackage{nopageno}

\usepackage{times}
\usepackage{epsfig}
\usepackage{graphicx}
\usepackage{amsmath}
\usepackage{amssymb}
\usepackage{textcomp}
\usepackage{amsmath}

\usepackage{multirow}
\usepackage{graphicx}
\usepackage{subcaption}
\usepackage{stfloats}
\newcommand{\squeezeup}{\vspace{-5mm}}
\newcommand{\smallsqueezeup}{\vspace{-2mm}}
\usepackage{enumitem}
\setlist[itemize]{noitemsep, topsep=0pt}

\usepackage[pagebackref=true,breaklinks=true,colorlinks,bookmarks=false]{hyperref}

\def\cvprPaperID{97} 
\def\confYear{CVPR 2021}
\begin{document}

\title{\ Physically Inspired Dense Fusion Networks for Relighting}


\author{Amirsaeed Yazdani\\
{\tt\small amiryazdani@psu.edu}
\and
Tiantong Guo\\
{\tt\small tiantong@ieee.org}
\and
Vishal Monga\\
{\tt\small vmonga@engr.psu.edu}
}


%

\maketitle
\squeezeup
\squeezeup
\squeezeup
\begin{abstract}
\squeezeup
\textit{Image relighting has emerged as a problem of significant research interest inspired by augmented reality applications.\ Physics-based traditional methods, as well as black box deep learning models, have been developed. The existing deep networks have exploited training to achieve a new state of the art; however, they may perform poorly when training is limited or does not represent problem phenomenology, such as the addition or removal of dense shadows.
We propose a model which enriches neural networks with physical insight.\ More precisely, our method generates the relighted image with new illumination settings via two different strategies and subsequently fuses them using a weight map ($w$).\ In the first strategy, our model predicts the  material reflectance parameters (\textbf{albedo}) and illumination/geometry parameters of the scene (\textbf{shading}) for the relit image (we refer to this strategy as \textbf{intrinsic image decomposition} (\textbf{IID})). The second strategy is solely based on the black box approach, where the model optimizes its weights based on the ground-truth images and the loss terms in the training stage and generates the relit output directly (we refer to this strategy as \textbf{direct}). While our proposed method applies to both \textbf{one-to-one} and \textbf{any-to-any} relighting problems, for each case we introduce problem-specific components that enrich the model performance: 1) For one-to-one relighting we incorporate normal vectors of the surfaces in the scene to adjust gloss and shadows accordingly in the image.\ 2) For any-to-any relighting, we propose an additional multiscale block to the architecture to enhance feature extraction. Experimental results on the VIDIT 2020 and the VIDIT 2021 dataset (used in the NTIRE 2021 relighting challenge) reveals that our proposal can outperform many state-of-the-art methods in terms of well-known fidelity metrics and perceptual loss.}
\end{abstract}
\squeezeup
\section{Introduction}
Image enhancement problems have experienced significant recent research activity inspired by the proliferation of mobile devices and the availability of training data designed for particular enhancement goals. Image relighting, which is changing the illumination settings of an image, is one of these applications that has attracted significant attention. Another important reason for this growth is the development of \textbf{augmented reality} (\textbf{AR}), \textbf{\textbf{virtual reality}} (VR) based services such as online shopping, online teaching, and games, where the gloss and shadows of the scene should be adjusted based on the change in direction of light and its color temperature. On the other hand, controlling the light source in the level of the photography skills of an amateur user is not trivial, which in turn necessitates the development of techniques for relighting. \\
From a physical viewpoint, the illumination of an image depends on many factors including the material reflectance property, geometry of the objects in the image, and the number of light sources. For a given light source ($L_{\omega_{i}}$) and with Lambertian reflectance assumption, the image formation follows the rendering rule \cite{34,7}:
\begin{equation}
    L_{\omega_{o}}=\int_{\omega_{i}\in\Omega_{o}}f(\omega_{i},\omega_{o})L_{\omega_{i}}< n,\omega_{i}>d\omega_{i}\smallsqueezeup
    \label{rendering}
\end{equation}
Here $\omega_{i}$ and $\omega_{o}$ denote the input and output light direction relative to the surface normal $n$. $L_{\omega_{i}}$ and $L_{\omega_{o}}$ are the incident and reflected lights, and $f(.,.)$ is the bidirectional reflectance distribution function (BRDF) and $< n,\omega_{i}>$ is the attenuation factor. This equation is usually simplified by assuming: $A=f(\omega_{i},\omega_{o})$ (constant) and $ S=\int_{\omega_{i}\in\Omega_{o}}L_{\omega_{i}}< n,\omega_{i}>d\omega_{i}$.
Where $A$ denotes albedo and preserves the reflectance properties of the objects and $S$ denotes shading, which holds the illumination properties of the image. The simplified equation is often used as the rendering rule as the original equation is computationally complex. 
\\
Deep learning methods have achieved state-of-the-art results for a vast variety of imaging inverse problems. High dynamic range (HDR) imaging algorithms \cite{28,29} focus on increasing the local contrast of a low dynamic range image. Dehazing algorithms \cite{30,31}  seek for removing the haze artifacts caused by floating particles in the atmosphere. Shadow removal \cite{33} and light enhancement \cite{32} methods focus on enhancing the lighting and removing the artifacts in the image with the existing light source. While all the aforementioned methods deal with adjusting the parameters affected by the lighting of the image, they don't manipulate actual illumination parameters and can not deal with complexities of relighting. Therefore, we are still in the early stages of relighting research. Existing algorithms focus on particular objects such as portraits or faces \cite{7,12,6}, hence lacking the versatility to generalize to other classes of objects (e.g buildings). Deep learning methods \cite{40,41,56} for relighting are versatile; however, they show poor performance in extreme cases of shadow removal/addition as they often ignore the physics of the problem. \\
\textbf{Our central contribution} is to generate relighted images with new illumination settings via two different strategies and subsequently fuse them using a weight map ($w$)
\begin{itemize}
    \item In the first strategy, the model predicts the  material reflectance parameters (\textbf{albedo}) and illumination/geometry parameters of the scene (\textbf{shading}) for the relit image and constructs the relit image based on the simplified rendering rule Eq. \ref{rendering}. (we refer to this strategy as \textbf{intrinsic image decomposition} (\textbf{IID}).)
    \item The second strategy follows a black box approach, where the model optimizes its weights based on the ground-truth images and the loss terms in the training stage and generates the relit output directly (we refer to this strategy as \textbf{direct}).
\end{itemize}
Our proposed method exploits insights from two different sides of the literature. Moreover, since both approaches have a shared encoder, owing to the virtue of joint optimization, they can benefit the shared features the other one induces the encoder to extract as well. In this work, we are addressing the problem of relighting under two categories: 
1) \textbf{One-to-one}: The objective is to change the color temperature and angle of the light source (referred to as illumination parameters) from one specific setting to another one. 
2) \textbf{Any-to-any}: The illumination parameters should change from an arbitrary setting according to the illumination of a given guide image. While our proposed method applies to both \textit{one-to-one} and \textit{any-to-any} relighting problems, for each case, we {\textbf {propose specific innovations}} that enrich the model performance:
\begin{itemize}
    \item  For one-to-one relighting we incorporate normal vectors of the surfaces in the scene to adjust gloss and shadows accordingly in the image. This in particular helps boost the performance of the neural network model in the cases where the complicated geometry of the scene requires removing dense shadows or adding shadows to highly glossed regions. We refer to our network for one-to-one relighting as \textbf{One-to-one Intrinsic Decomposition-Direct RelightNet} (\textbf{OIDDR-Net)}.
    \item For any-to-any relighting, we propose an additional multiscale block to the architecture to enhance feature extraction. This block benefits from analyzing the input RGB image (and depth map) in three different dimension levels. Using dense residual blocks and residual global blocks in each level, it provides multiscale features for the subsequent layers. We refer to our network for any-to-any relighting as \textbf{Any-to-any Multiscale Intrinsic-Direct RelightNet} (\textbf{AMIDR-Net)}.
\end{itemize}
Our experimental results on VIDIT 2020/2021 dataset prove that our proposed method can outperform state-of-the-art in terms of fidelity metrics and perceptual loss. \textbf{Our OIDDR-Net ranked second and AMIDR-Net ranked among top five teams in NTIRE 2021 depth guided image relighting challenge \cite{59}}.


\section{Related Works}
The existing works in the area of image relighting can generally be divided into two groups of deep learning-based methods and conventional image processing methods. In the line of conventional methods and starting with models proposed for inverse rendering \cite{6,25} and shape estimation \cite{5}, there have been relighting works based on decomposing the image into its reflectance, illumination, and geometry components. Duch\^{e}ne \emph{et al}. \cite{35} utilize a set of outdoor multiview scenes along with the sunlight direction to achieve albedo and shading decomposition for relighting. Wen \emph{et al.} \cite{36} develop a technique in which the estimated radiance environment maps, along with spherical harmonics, are used for face relighting. Other algorithms \cite{37,38,39} treat relighting as approximating the light transport function of the scene to generate the new illumination settings using the input lighting parameters.  While these methods construct physically realistic models for relighting, they rely on explicit illumination parameters of the scene or multiview datasets. This is considered a bottleneck for them.\\   
While conventional methods mostly focus on physical aspects of the problem, deep learning-based algorithms rely on the capability of neural networks, along with typically large training data set, in developing a mapping function between two image domains. Methods proposed in \cite{40,41} view the relighting as an image-to-image translation problem and make use of Generative Adversarial Networks (GANs) for image relighting. Xu \emph{et al.} \cite{42} use five images to manipulate the illumination under predefined light direction. Inspired by inverse rendering, numerous works incorporate neural networks to estimate image illumination and geometry parameters. Yu \emph{et al}. \cite{21,3} view relighting as a fruit of regressing the albedo, shading, and light coefficients of the input RGB image using fully convolutional networks. Face relighting methods \cite{12,7,15} combine the capabilities of neural networks and the physics of relighting, which is customized for face images. Although introducing neural networks has shown promising results for the relighting problem, the appropriate dataset yet seems to be a challenging aspect. In the past years, IIW \cite{44} and MIP SINTEL \cite{43} have been introduced for intrinsic decomposition and  optical flow analysis, respectively. More recently, Helou \emph{et al.} proposed a virtual image dataset for illumination transfer \cite{13}, which formulates relighting into one-to-one and any-to-any problems. Along the line of scene relighting and illumination estimation challenge in AIM 2020 \cite{1}, Puthessery \emph{et al.} \cite{14} propose a U-net \cite{45} model for one-to-one relighting where Discrete Wavelet Transform (DWT) and Inverse Discrete Wavelet Transform (IDWT) are attached to downsampling and upsampling layers, respectively. SA-AE \cite{46} also develops a U-Net-based architecture for any-to-any relighting, where two auxiliary networks are incorporated for estimating the lighting of the guide image and providing lighting features to the decoder.

\begin{figure*}
    \centering
    \includegraphics[width=.9 \linewidth]{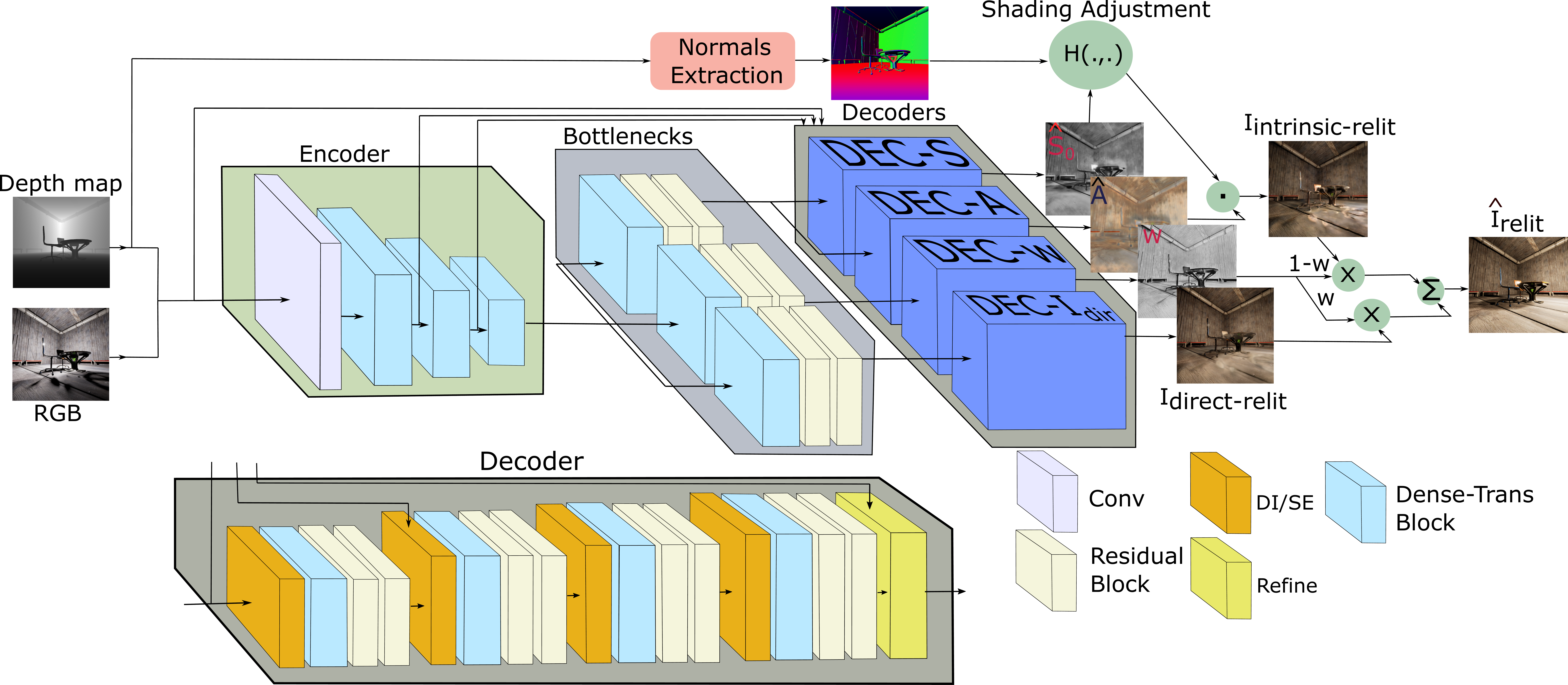}
    \smallsqueezeup
    \caption{Our proposed OIDDR-Net architecture.}
    \label{fig:one_to_one}
\end{figure*}
\begin{figure}
    \centering
    \includegraphics[width=0.85 \linewidth]{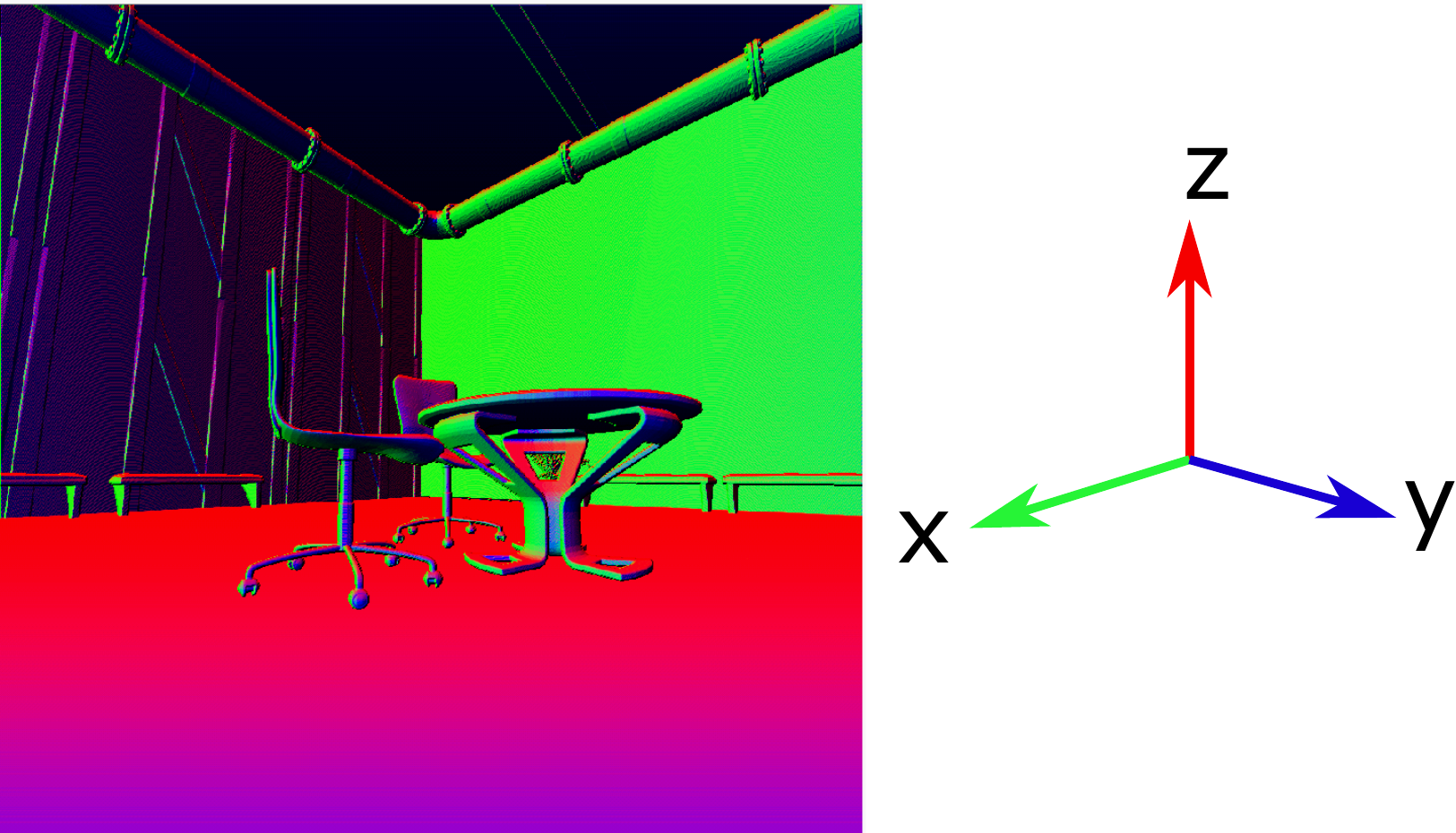}
    \smallsqueezeup
    \caption{The visualization of the normal vectors in an example image. Each of RGB channels corresponds to x-y, y-z, and x-z planes, respectively.}\squeezeup
    \label{fig:normal}
\end{figure}
\smallsqueezeup
\begin{figure*}
    \centering
    \includegraphics[width=0.85 \linewidth]{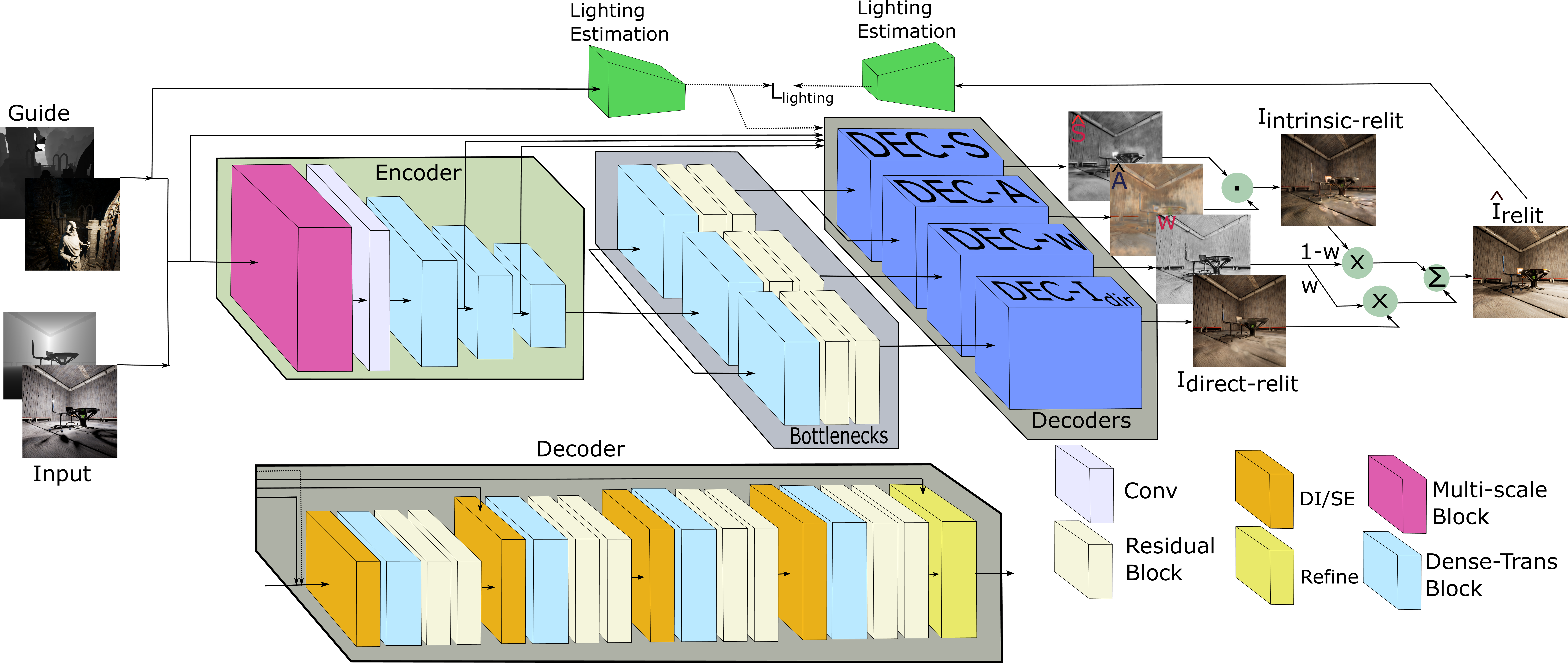}
    \caption{Our proposed AMIDR-Net architecture.}
    \label{fig:any_to_any}
\end{figure*}
\begin{figure}
    \centering
    \includegraphics[width=1\linewidth]{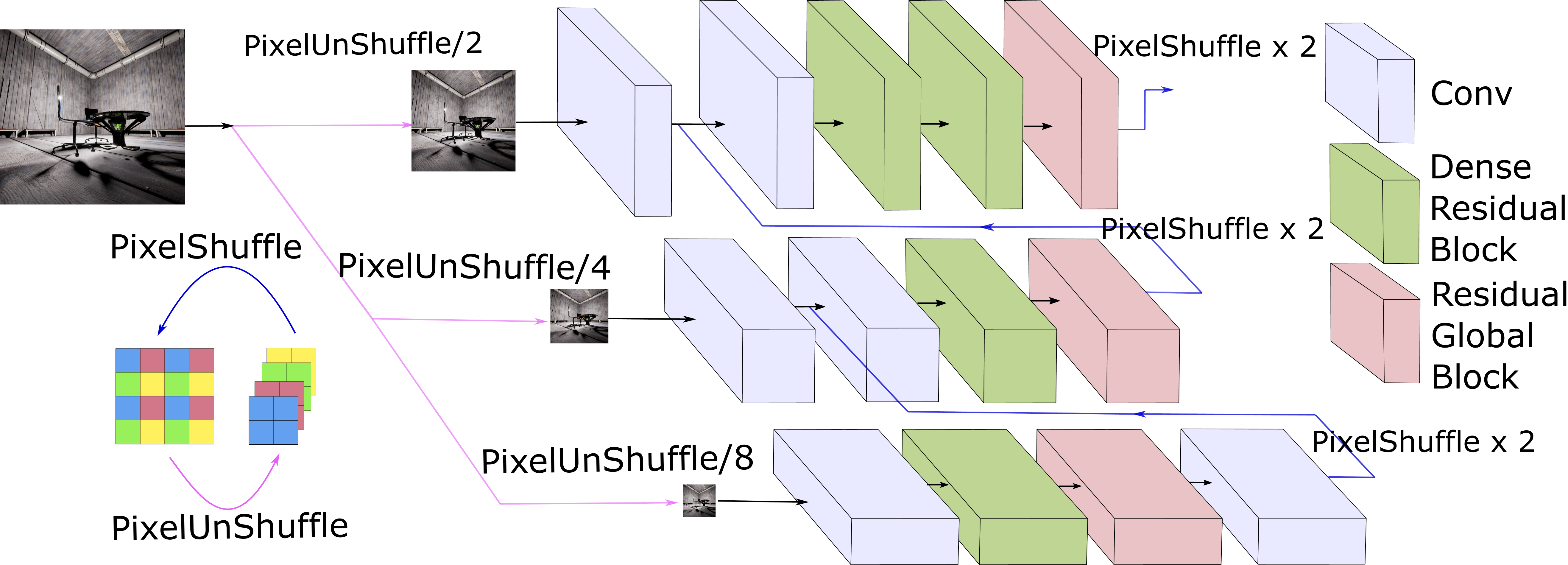}
    \squeezeup
    \caption{The multiscale block.}
    \label{fig:multiscale}
    \smallsqueezeup
\end{figure}
\begin{figure}
\setlength\belowcaptionskip{-20pt}
    \centering
    \includegraphics[width=1\linewidth]{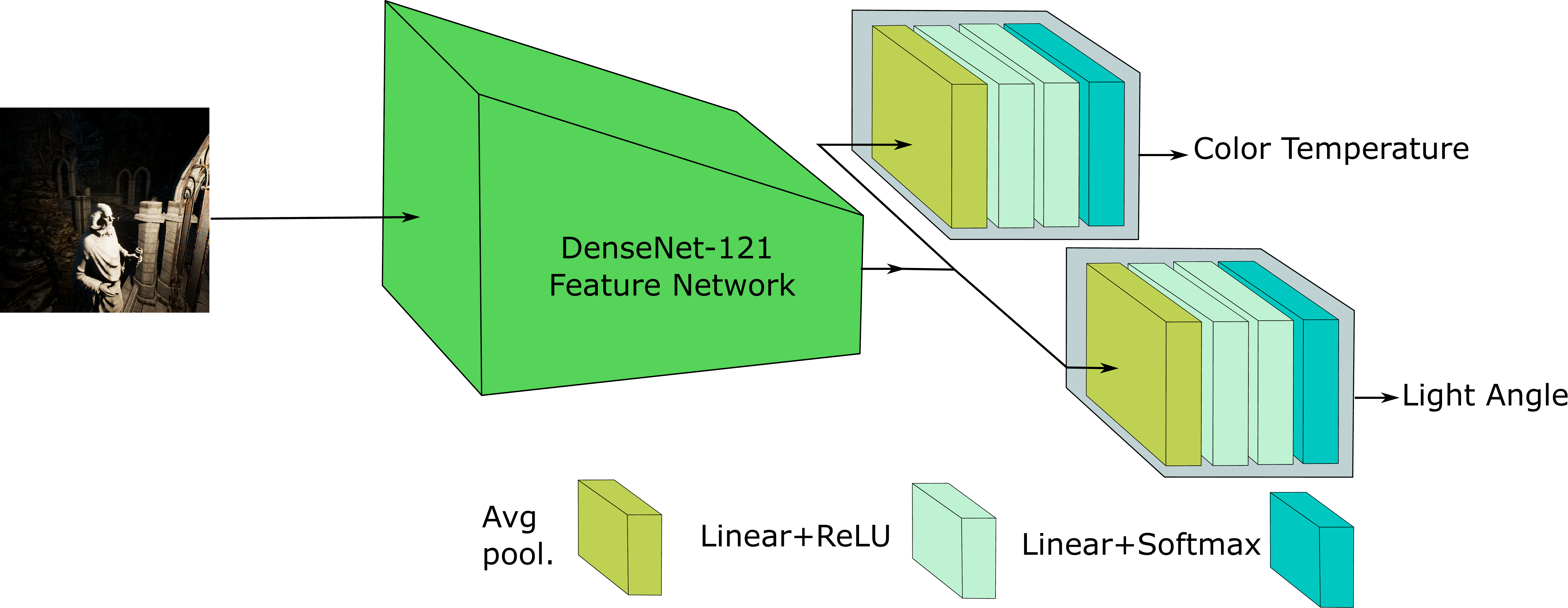}
    \squeezeup
    \caption{The lighting estimation network used for illumination regularization and extraction of illumination features from the guide image.}
    \label{fig:light_estimation}
\end{figure}
\section{Proposed Method}
\smallsqueezeup
\subsection{Fusion Strategy}
\smallsqueezeup
As mentioned earlier, our model is based on two approaches to the relighting problem:\\
\textbf{1) Intrinsic Image Decomposition} (\textbf{IID}): From the physical standpoint, every light image can be decomposed into two main parameters \cite{24,25}: \textbf{albedo} and \textbf{shading}. Albedo, which is the light independent parameter of the image, preserves the reflectance property of the material in the scene, while shading holds properties corresponding to illumination and the geometry of the image. Based on this, the relit image can be expressed as: $ I_{\text{intinsic-relit}}=\hat{A} \odot \hat{S}$.
 Here, $\hat{A}$ and $\hat{S}$ denote estimated albedo and shading, respectively. $\odot$ is the element wise product operation. This method has been shown as an effective way to relight the image scene using the input  RGB image (and depth map) \cite{18,17,8,21,3}. Following this approach, the model is guided toward a systematic way of learning to relight by which it can distinguish between features associated with material reflectance property and features for the illumination and geometry of the scene.\\
\textbf{2) Direct Relighting} (\textbf{DR}): In addition to the intrinsic decomposition of the images, we also follow the end-to-end learning method as in state of the art \cite{10,9,14} for learning a mapping function between the two lighting settings: $ f(I)=I_{\text{direct-relit}}$.
 Where $f$ denotes the mapping function learned by neural network model. This way the model, in addition to the physically inspired insight, constructs an auxiliary insight that complements the other one in terms of the extracted discriminative features. \\
Next, we generate a spatially varying weight map ($w$) to fuse the estimates:
\smallsqueezeup
\begin{equation}
    \hat{I}_{\textit{relit}}=wI_{\textit{direct-relit}}+(1-w)I_{\textit{intrinsic-relit}}\smallsqueezeup
\label{eq:fusion}
\end{equation}
This fusion strategy helps the model benefit from both aspects of the problem simultaneously. Furthermore, owing to the usage of a couple of shared structures and the virtue of joint optimization, each approach aids the other one through the insight it lends to the model.

\begin{table*}[]\centering\caption{Encoder Structure for AMIDR-Net. ($^{*}$OIDDR-Net doesn't have the multiscale block and guide inputs.)}\squeezeup\label{tb:encoder}\medskip
\resizebox{0.85\textwidth}{!}{
\begin{tabular}{l|ccccc}\hline
     &Multi-Scale block & Base  & Dense-Trans.1 & Dense-Trans.2 & Dense-Trans.3 \\\hline
Input &$[\text{Input image},\text{Guide image}^*,~\text{Input depth map},~\text{Guide depth map}^*]$ &Multi-scale Output& Base&  Dense-Trans.1 &Dense-Trans.2 \\
\multirow{4}{*}{structure}&\multirow{4}{*}{See Fig. \ref{fig:multiscale}}& \multirow{2}{*}{$\begin{bmatrix}7\times7~\text{conv.}\\ \\3\times3~\text{max-pool}\end{bmatrix}$} &

$\begin{bmatrix}1\times1~\text{conv.}\\3\times3~\text{conv.}\end{bmatrix}\times6$ &

$\begin{bmatrix}1\times1~\text{conv.}\\3\times3~\text{conv.}\end{bmatrix}\times12$&  

$\begin{bmatrix}1\times1~\text{conv.}\\3\times3~\text{conv.}\end{bmatrix}\times24$\\
&&&$\begin{bmatrix}1\times1~\text{conv.}\\2\times2~\text{avg-pool}\end{bmatrix}$&$\begin{bmatrix}1\times1~\text{conv.}\\2\times2~\text{avg-pool}\end{bmatrix}$&
$\begin{bmatrix}1\times1~\text{conv.}\\2\times2~\text{avg-pool}\end{bmatrix}$\\

Output & $384\times384\times8$&$96\times96\times64$ & $48\times48\times128$ & $24\times24\times256$ & $24\times24\times512$ \\\hline
\end{tabular}}
\end{table*}
\subsection{Network Architecture}\smallsqueezeup
OIDDR-Net (Fig. \ref{fig:one_to_one}) and AMIDR-Net (Fig. \ref{fig:any_to_any}) are similar in terms of the general architecture which is inspired by U-Net \cite{45,4}. An encoder is shared by three bottlenecks followed by four decoders. Two decoders designed for the IID strategy share a bottleneck, while the other two decoders designated for direct strategy and generating the weight map ($w$) are fed by a bottleneck, each. The details of these components are as follows:\\
\textbf{1) Encoder}: The encoder construction (Table \ref{tb:encoder}) follows DenseNet-121 \cite{22} feature extraction part, which is originally proposed for classification tasks. It consists of a feature extraction part followed by classification layers. We borrow the input convolutional layer, the first three dense layers, and their following transitional blocks in the feature extraction part. The main advantage of using these pretrained layers for our model is that since they're trained over ImageNet dataset \cite{47}, they provide our model an initial representation capability. This in turn helps for the faster convergence in the training. It is worth mentioning that the first convolutional layer in DenseNet accepts three channels RGB images and inputs to OIDDR-Net and AMIDR-Net are of 4 and 8 channels, respectively. To address this, for OIDDR-Net, we modify the convolutional block by keeping the first three channels and initializing the fourth one as a grayscale transformation of the other three. For AMIDR-Net, we substitute it by an eight channel convolutional block initialized randomly.\\
\textbf{2) Bottlenecks}: There are three bottlenecks consisting of a dense transitional block and two residual blocks (see Table \ref{tb:encoder},\ref{tb:decoder} for details of these blocks). The main function of bottlenecks is to connect the encoder and decoders by compiling the extracted features of the encoder based on the characteristics of the decoders. So, Decoder-A and Decoder-S share a bottleneck as they both contribute on $I_{\text{intrinsic-relit}}$.\\
\textbf{3) Decoders}: Our network features four decoders for predicting the components: albedo ($\hat{A}$), shading ($\hat{S}$), weight map ($w$), and directly relit image ($I_{\text{direct-relit}}$). Table \ref{tb:decoder} details the structure of decoders. Each decoder includes four levels of cascaded attention module (Squeeze and excitation \cite{48} or dilation inception modules \cite{30}), a dense transitional block, and two residual blocks. It means that there is an analogy between decoders' structure and the encoder except that the channel attention modules are incorporated midway. This in particular helps each decoder give weight to feature maps based on its functionality, while benefiting the shared encoder. Meanwhile, skip connections from the encoder assist the decoders in reconstructing the scene as in U-net.\\
\textbf{4) Lighting Estimation Network}: As our main objective in this work is to change the illumination of the images, we need custom blocks for extracting illumination features. Moreover, it is worth mentioning that neural networks usually need either ground-truth or custom loss terms in order to extract our expected features. To this end, we train a lighting estimation network to predict the light angle and color temperature of the images using the training set for the any-to-any problem. We make use of the pretrained feature extraction part of this network (Fig. \ref{fig:light_estimation}) to compute a perceptual loss for comparing the illumination features of the relit output. Furthermore, in AMIDR-Net (Fig. \ref{fig:any_to_any}), this network is incorporated for feeding the decoders with the illumination features of the guide image.
\begin{table*}[]\centering\caption{Decoder Structure. (for OIDDR-Net the input to the decoder doesn't include lighting estimation outputs.) C depends on the functionality of the decoder.}\squeezeup\label{tb:decoder}\medskip
\resizebox{0.85\textwidth}{!}{
\begin{tabular}{l|ccccc}\hline
&Dense-Trans.5&Res.5&Dense-Trans.6&Res.6&Dense-Trans.7\\ \hline
Input&$[\text{bottleneck output},~\text{Dense.Trans.2},\text{Lighting Estimation}]$&Dense-Trans.5&$[\text{Trans.1},~\text{Res.5}]$&Dense-Trans.6&Res.6\\
\multirow{4}{*}{structure}&

$\begin{bmatrix}\text{SE/Dilation (R=16)} \\\text{batch norm} \\ 3\times3~\text{conv.}\end{bmatrix}\times7$&

\multirow{2}{*}{$\begin{bmatrix}3\times3~\text{conv.}\\ \\3\times3~\text{conv.}\end{bmatrix}\times2$}&

$\begin{bmatrix}\text{SE/Dilation (R=16)} \\\text{batch norm}\end{bmatrix}\times7$&

\multirow{2}{*}{$\begin{bmatrix}3\times3~\text{conv.}\\ \\3\times3~\text{conv.}\end{bmatrix}\times2$}&
$\begin{bmatrix}\text{batch norm} \\ 3\times3~\text{conv.}\end{bmatrix}\times7$\\

&$\begin{bmatrix}1\times1~\text{conv.}\\~\text{upsample 2}\end{bmatrix}$&&$\begin{bmatrix}1\times1~\text{conv.}\\~\text{upsample 2}\end{bmatrix}$&&$\begin{bmatrix}1\times1~\text{conv.}\\~\text{upsample 2}\end{bmatrix}$\\
output&$48\times48\times128$&$48\times48\times128$&$96\times96\times64$&$96\times96\times64$&$192\times192\times32$\\ \hline \hline 
&Res.7&Dense-Trans.8&Res.8&Refine.9&Refine.10\\ \hline
Input&Dense-Trans.7&Res.7&Dense-Trans.8&$[\text{Input},~\text{Res.8}]$&Refine.9\\
\multirow{4}{*}{structure}&\multirow{2}{*}{$\begin{bmatrix}3\times3~\text{conv.}\\ \\3\times3~\text{conv.}\end{bmatrix}\times2$}&$\begin{bmatrix}\text{batch norm} \\ 3\times3~\text{conv.}\end{bmatrix}\times7$&\multirow{2}{*}{$\begin{bmatrix}3\times3~\text{conv.}\\ \\3\times3~\text{conv.}\end{bmatrix}\times2$}&\multirow{2}{*}{$\begin{bmatrix}\text{SE/Dilation (R=3)} \\\text{batch norm} \\ 3\times3~\text{conv.}\end{bmatrix}$}&\multirow{2}{*}{$\begin{bmatrix}32\times32~\text{avg-pool} \\1\times1~\text{conv.}\\\text{upsample}\end{bmatrix}$}\\
&&$\begin{bmatrix}1\times1~\text{conv.}\\~\text{upsample 2}\end{bmatrix}$&&&\\
output&$192\times192\times32$&$384\times384\times16$&$384\times384\times16$&$384\times384\times20$&$384\times384\times1$\\ \hline \hline
&Refine.11&Refine.12&Refine.13&Output.14&\\
\hline
Input&Refine.9&Refine.9&Refine.9&$[\text{Refine9.10.11.12.13}]$&\\ 
structure&$\begin{bmatrix}16\times16~\text{avg-pool} \\1\times1~\text{conv.}\\\text{upsample}\end{bmatrix}$&$\begin{bmatrix}8\times8~\text{avg-pool} \\1\times1~\text{conv.}\\\text{upsample}\end{bmatrix}$&$\begin{bmatrix}4\times4~\text{avg-pool} \\1\times1~\text{conv.}\\\text{upsample}\end{bmatrix}$&$3\times3~\text{conv.}$\\ Output&$384\times384\times1$&$384\times384\times1$&$384\times384\times1$&$384\times384\times C$\\ \hline
\end{tabular}}
\smallsqueezeup
\smallsqueezeup
\smallsqueezeup
\end{table*}
\subsection{Exploiting Normals for One-to-One Relighting}
While the model is guided toward learning a physics-based solution for relighting, it may not necessarily be successful in changing the lighting parameters of a given scene. This could be due to the complicated geometry of the scene, which causes the presence of dense shadows needed to be removed or the presence of highly glossed objects needed to be shadowed. This is a challenging aspect of relighting for a neural network model, as neural networks usually fail in regressing outputs that lie on one side of the extreme since their share in training data is typically small. Therefore, the model in order to keep its generalization over the whole data distribution would typically show artifacts on these extreme cases. To address this issue specifically, for the case of one-to-one relighting, we propose to incorporate the information associated with the normal vectors of the surfaces present in the scene. It is shown that the shading of an image can be derived as a nonlinear function of 9-dimensional spherical harmonics coefficients and the normal vectors \cite{7,8,24,26}. The normal vectors of a scene indicate the orientation of pixels associated with each surface in the image. Fig. \ref{fig:normal} shows an example in which the colors red, green, and blue indicate surfaces parallel to x-y, y-z, and x-z plane, respectively. Since our model learns to predict the shading directly from the information provided during the training stage in the form of ground-truth, instead of carrying out the non-linear calculations, we incorporate normal vectors into the problem as weight adjustments.\ More precisely, in the case of one-to-one relighting in which we know the target lighting direction, the normal vectors are used as adjustment weights for the surfaces facing toward or against the light direction: $ \hat{S}=H(n_{\textit{light-dir}},\hat{S_{0}})$.
 Where $H$, $n_{\textit{light-dir}}$, and $\hat{S_{0}}$ are the linear adjustment function, the normal vector component corresponding to the light direction of the target, and the shading output by the model, respectively.
\subsection{Any-to-Any Relighting and Multiscale Features}
In any-to-any relighting there is no meaningful pixel-wise correspondence between the guide and input RGB image/depth map. Therefore unlike one-to-one relighting, training the network on image patches is not feasible. On the other hand, training the network on whole images limits the representation power of the model as the model may not necessarily extract features from lower fields of view. To prevent that, we equip AMIDR-Net with a multiscale feature extraction block \cite{23}. Fig. \ref{fig:multiscale} shows the details of this block. Using PixelUnShuffle operations, it downsamples the input to three different levels. In each level dense residual and global residual blocks extract the features. Subsequently, the extracted feature maps are \textbf{customly} upsampled and fed to the next level using PixelShuffle operation. Finally, in the highest level the multiscale feature maps are processed and fed to the main pipeline. The main advantage of using this multiscale block over the ones, which make use of traditional upsampling modules, is how it guides the network to learn the upsampling while optimizing the feature maps. Simply put, the model learns how to extract patches from the input while keeping the correspondence between the feature maps from the guide and input.
\subsection{Customized Loss Function}
To train our model so that every part of it functions based on our expectation, we need to define custom loss terms for each part. Our overall loss function is as follows:\smallsqueezeup
\small
\begin{multline}
    \mathcal{L}=\mathcal{L}_{\textit{total}}+\lambda_{1}\mathcal{L}_{IID}+\lambda_{2}\mathcal{L}_{\textit{direct}}+\lambda_{3}\mathcal{L}_{\textit{SSIM}}+\lambda_{4}\mathcal{L}_{\textit{lighting}}
\end{multline}
\squeezeup
\begin{equation}
    \mathcal{L}_{\textit{total}}=||\hat{I}_{\textit{relit}}-Y_{\textit{relit}}||_{2}^{2}
\end{equation}
\begin{equation}
  \mathcal{L}_{IID}=||\hat{A}\odot \hat{S}-Y_{\textit{relit}}||_{2}^{2}+||\hat{A}-A||_{2}^{2}+||\hat{S}-S||_{2}^{2}
\end{equation}
\begin{equation}
  \mathcal{L}_{\textit{direct}}=||I_{\textit{direct-relit}}-Y_{\textit{relit}}||_{2}^{2}
\end{equation}
\begin{equation}
  \mathcal{L}_{SSIM}=1-SSIM(\hat{I}_{\textit{relit}},Y_{\textit{relit}})\smallsqueezeup
\end{equation}
\begin{multline}
\setlength{\abovedisplayskip}{0pt}
\setlength{\belowdisplayskip}{3pt}
 \mathcal{L}_{\textit{lighting}}=||g(\hat{I}_{\textit{relit}})-g(Y_{\textit{relit}})||_{2}^{2}\\\color{blue}-\sum_{i=1}^{8}Y_{\textit{dir-guide}}^{i}log(\hat{Y}_{\textit{dir}}^{i})-\sum_{j=1}^{5}Y_{\textit{color-guide}}^{j}log(\hat{Y}_{\textit{color}}^{j})
 \label{eq:lighting}
 \end{multline}
 \normalsize
 \color{black}
 Where $\mathcal{L}_{\textit{total}}$, $\mathcal{L}_{IID}$, and $\mathcal{L}_{\textit{direct}}$ are terms to ensure the decoder outputs $I_{\textit{intrinsic-relit}}$, $I_{\textit{direct-relit}}$, and $\hat{I}_{\textit{relit}}$ match the ground-truth $Y_{\textit{relit}}$. Of note, $\hat{I}_{\textit{relit}}$ is the fused output (eq. \ref{eq:fusion}). To help the model predict physically feasible estimates for albedo and shading, we use a pretrained intrinsic decomposition network \cite{3}, which is trained on SINTEL dataset \cite{43}, to generate pseudo ground-truths $A$ and $S$. $\mathcal{L}_{SSIM}$ is used to maximize the structural similarity index (SSIM) of the relit output and ground-truth. We also define $\mathcal{L}_{\textit{lighting}}$ to minimize the difference between the relit output and the ground-truth in terms of illumination parameters. In eq. \ref{eq:lighting}, in the first term, the intermediate features generated by lighting estimation network (denoted by $g$) are compared for the relit output and ground-truth. The second and third terms (blue) are specifically incorporated for AMIDR-Net, where we minimize the negative log-likelihood of the light direction and color temperature in the relit output ($\hat{Y}_{\textit{dir}}$ and $\hat{Y}_{\textit{color}}$) based on guide image parameters ($Y_{\textit{dir-guide}}$ and $Y_{\textit{color-guide}}$). $\lambda_{1}$, $\lambda_{2}$, $\lambda_{3}$, and $\lambda_{4}$ are hyperparameters adjusting the contribution of each term in the overall loss term.
 \section{Implementation Details}
 \subsection{Dataset}
 We use VIDIT dataset \cite{13} generated by the Unreal gaming engine \cite{58} and consisting of two subsets for one-to-one and any-to-any relighting.\\ 
\textbf{One-to-One Relighting}: VIDIT'21 training set for one-to-one relighting, provides 300 $1024\times1024$ images with one particular light direction and color temperature  and their corresponding ground-truth with the target particular illumination settings. Additionally, in VIDIT'21, unlike VIDIT'20, depth maps are provided for each image. The validation set includes 45 samples. We augment\footnote{ Please find implementation details and results at: \href{https://github.com/yazdaniamir38/Depth-guided-Image-Relighting}{\underline{github/Relighting}}} the training set to a set with about 37000 samples by 1) cropping 256$\times$256 patches.\ 2) Resizing the whole images to $256\times256$. 3) Rotating the patches and resized images slightly with small angles (0-12 degrees). We don't incorporate flipping or rotation with large angles as the light direction in this problem should be fixed across the training samples.\\
 \textbf{Any-to-Any Relighting}: The diversity across the training set is higher in the case of any-to-any relighting. For training VIDIT provides 12000 samples consisting of 300 scenes with a combination of 8 different light angles and 5 different color temperatures (40 for each scene) and 1 depth map for each scene. The validation set includes 90 images. As mentioned earlier, we cannot crop patches in this case; however, training on whole $1024\times1024$ images is not possible due to memory limits. Therefore, we resize the images to $384\times384$. We create the training set following two steps: 1) For each sample in the set,  we randomly choose three different guide samples. The guide samples, obviously, are not from the same scene as the original sample. 2) For each of the chosen guide images, we find the version of original scene having the same illumination setting as the guide image. This leads to a training set with 36000 samples.
 \subsection{Training}
We use Adam optimizer \cite{49} with an initial learning rate of $10^{-4}$, which decreases by a rate of 0.7 every 10 epochs. Owing to pretrained weights of DenseNet and incorporation of pseudo ground-truths, both OIDDR-Net and AMIDR-Net show fast convergence (optimally 20 epochs), but we train the models for 25 epochs (with batch sizes of 8 and 2, respectively) to ensure the complete stability of them. $\lambda_{1}$, $\lambda_{2}$, $\lambda_{3}$, and $\lambda_{4}$ are set to 0.4, 0.4, 0.8, and 0.03, respectively using cross validation \cite{50}.
\subsection{Testing}
 \textbf{One-to-One Relighting}: We observe that our model shows better performance by the following ensemble method:
i) $1024\times1024$ RGB image and depth map are fed to the model. ii) $384\times384$ RGB image and depth map are input to the model. The output is fed to a \textbf{bicubic} interpolation module (implemented in pytorch \cite{51}) and scaled to the original size. The final output is the average of the two estimates.\\
\textbf{Any-to-Any Relighting}:
In order to get the best performance of AMIDR-Net during the test phase, we resize the input to $384\times384$ so the model has the same observation as in training. The outputs then will be upsampled using bicubic interpolation.\smallsqueezeup  
\section{Experimental Results}
In this section we present experimental results of our proposed OIDDR-Net and AMIDR-Net.\ We provide ablation studies to show the effect of loss terms and novel network components.\ We also compare our models with state of the art. Our evaluation metrics are Peak Signal-to-Noise Ratio (PSNR), Structural Similarity Index (SSIM) \cite{57}, Learned Perceptual Image Patch Similarity (LPIPS) \cite{52}, and Mean Perceptual Score (MPS) which is: $\hspace{15ex}   MPS=0.5(SSIM+1-LPIPS)$.\smallsqueezeup
\begin{table}[]
\small
\caption{One-to-one-VIDIT'21 validation's ablation study.}
\smallsqueezeup
\label{tb:ablation1}
    \centering
    \resizebox{.4\textwidth}{!}{
    \begin{tabular}{c|c|c|c|c}
    \hline
         Model&PSNR&SSIM&LPIPS&MPS  \\
         \hline
         OIDDR-Net& \textbf{18.39}&\textbf{0.6980}&\textbf{0.2591}&\textbf{0.7194}\\
         \hline
         w/o Normals&17.49&0.6805&0.2647&0.7079\\
         \hline
         w/o $L_{\textit{lighting}}$&17.59&0.6669&0.2741&0.6964\\
         \hline
    \end{tabular}}
    \label{tab:my_label}
    \smallsqueezeup
    \smallsqueezeup
\end{table}
\subsection{Ablation Study}
\textbf{Effect of Exploiting Normal Vectors and $L_{\textit{lighting}}$ on OIDDR-Net}:
To observe how incorporating normal vectors for adjusting the shading estimation of the network would affect its performance, we conduct an experiment on VIDIT'21 dataset through which we train OIDDR-Net with initial shading estimation. Additionally, to study the effect of comparing the illumination in network relit output to illumination in ground-truth (through $L_{\textit{lighting}}$), we train OIDDR-Net without applying $L_{\textit{lighting}}$ in training loss. Table \ref{tb:ablation1} shows the performance results of the three models. While both factors play an important role in minimizing the fidelity and perceptual loss, we can see how $L_{\textit{lighting}}$ contributes to structural similarity.\\
\textbf{Effect of Multiscale Feature Extraction and $L_{\textit{lighting}}$ on AMIDR-Net}: To see the importance of the multiscale block as well as $L_{\textit{lighting}}$, we train three models: 1) AMIDR-Net with full architecture and $L_{\textit{lighting}}$ being activated during training, 2) AMIDR-Net with the multiscale block removed from the architecture, and 3) AMIDR-Net trained with $L_{\textit{lighting}}$ dropped from the training loss. Table \ref{tb:ablation2} shows the results of our experiments on validation data, whereby one can infer the effect of different components on the network performance. The noticeable drop in SSIM, after removing the multiscale block, proves its impact in helping the network extract more discriminative features for constructing the structural similarity between the output and the ground-truth.
\begin{table}[]
\caption{Any-to-any-VIDIT'21 validation's ablation study.}
\smallsqueezeup
\label{tb:ablation2}
    \centering
\resizebox{.4\textwidth}{!}{    \begin{tabular}{c|c|c|c|c}
    \hline
         Model&PSNR&SSIM&LPIPS&MPS  \\
         \hline
         AMIDR-Net& \textbf{19.83}&\textbf{0.6940}&\textbf{0.3381}&\textbf{0.6779}\\
         \hline
         w/o the multiscale block&19.09&0.6685&0.3421&.6632\\
         \hline
         w/o $L_{\textit{lighting}}$&19.22&0.6721&0.3403&0.6659\\
         \hline
    \end{tabular}}
    \smallsqueezeup
\end{table}
\begin{figure*}[h]
    \centering
    \includegraphics[width=.8 \linewidth]{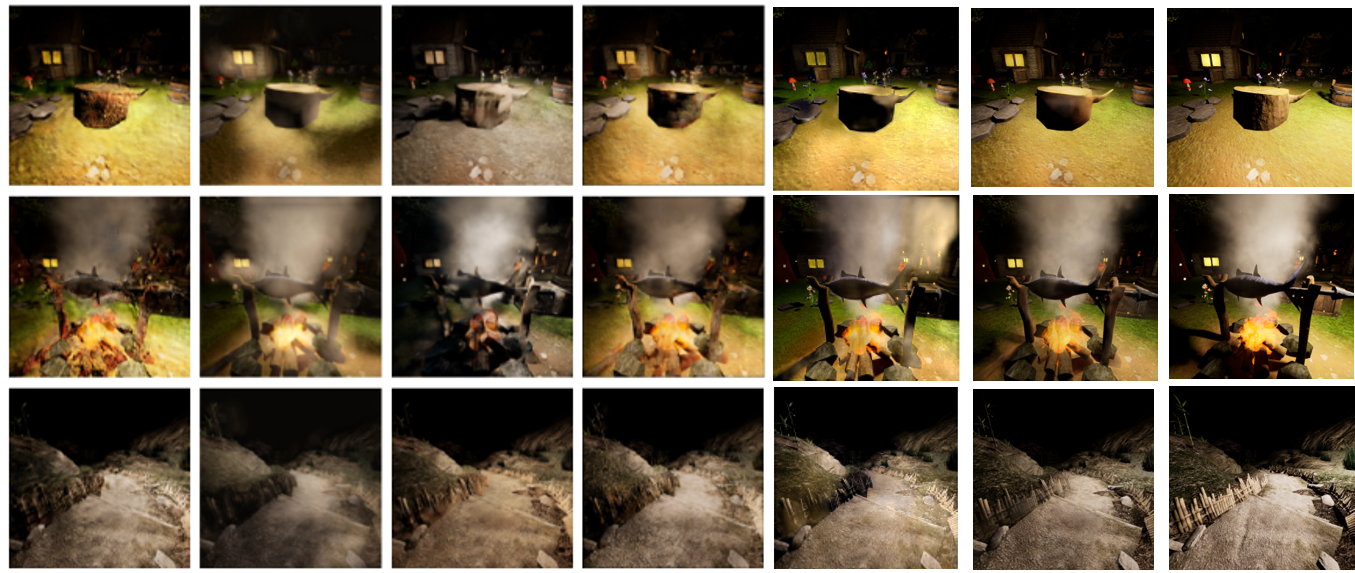}\smallsqueezeup\vspace{-1mm}
    \caption{Qualitative comparison between different methods on one-to-one relighting. From left to right: SRN \cite{53}, Dense-GridNet \cite{54}, DRN \cite{11}, DMSHN \cite{55}, WDRN \cite{14}, OIDDR-Net (ours) and ground-truth.  }\smallsqueezeup
    \label{fig:comp1}
    \smallsqueezeup
\end{figure*}
\begin{table}[]
\caption{Comparison with state of the arts for one-to-one relighting on VIDIT'20 validation set.}
\smallsqueezeup
\label{tb:comp1}\centering
\resizebox{.5\textwidth}{!}{
    \begin{tabular}{c|c|c|c|c|c}
    \hline
         Model&PSNR&SSIM&LPIPS&MPS&Runtime(s)  \\
         \hline
         OIDDR-Net (ours)& \textbf{17.62}&\textbf{0.6645}&\textbf{0.2733}&\textbf{0.6956}&0.53\\
         \hline
        WDRN \cite{14}&17.45&0.6642&0.2771&0.6935&0.05\\
         \hline
         DRN \cite{11}&17.59&0.596&0.440&0.578&0.5\\
         \hline
         DMSHN \cite{55}&17.20&0.5696&0.3712&0.5992&0.0058\\
         \hline
         SRN \cite{53}&16.94&0.5660&0.4319&0.5670&0.87\\
         \hline
         Dense-GridNet \cite{54}&16.67&0.2811&0.3691&0.9120&0.9326\\
         \hline
         Dong \emph{et al.} \cite{56}&17.14&0.6132&0.2764&0.6684&---\\
         \hline
    \end{tabular}}
    \squeezeup
\end{table}
\subsection{Comparison with State-of-the-art Methods}
\begin{figure*}
    \centering
    \includegraphics[width=.8 \linewidth]{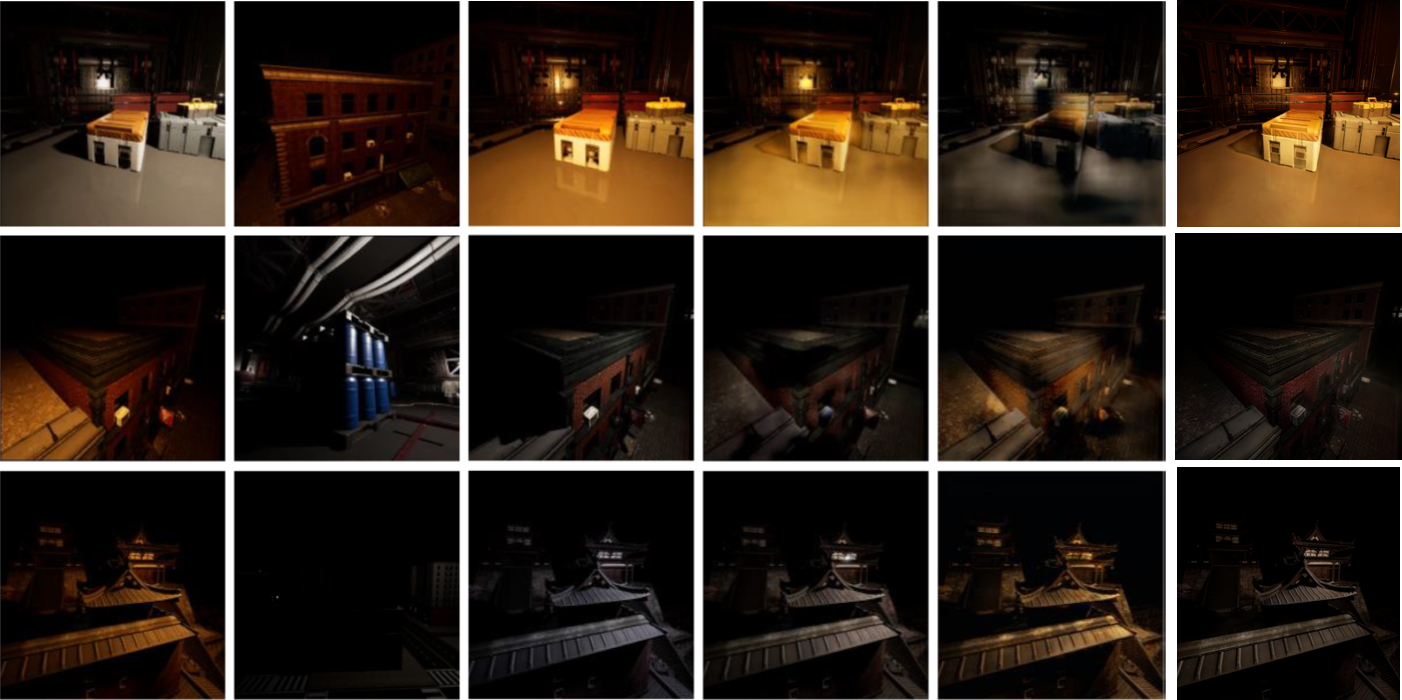}\smallsqueezeup
    \caption{Qualitative comparison between different methods. From left to right: input image, guide image, ground-truth, SA-AE \cite{10}, DPR \cite{15}, and AMIDR-Net (ours).}
    \label{fig:comp2}\squeezeup
\end{figure*}

All the existing works for image-based relighting (applicable to VIDIT) have been proposed for the dataset without depth information. Therefore, to have a fair comparison, we trained and evaluated our OIDDR-Net and AMIDR-Net on VIDIT'20 training and validation set.\\
\textbf{One-to-one Relighting}:
We modify our OIDDR-Net for accepting only the RGB image (so normals are not exploited) and train it over the training set. We compare our modified OIDDR-Net with existing methods in Table \ref{tb:comp1}. While \cite{53} and \cite{54} are proposed for deblurring and dehazing, all other methods have been proposed for the same exact problem and dataset. Table \ref{tb:comp1} shows that our OIDDR-Net outperforms state of the art w.r.t. all metrics as a result of fusing the power of neural networks and the physics of the problem. We can also qualitatively confirm this in Fig. \ref{fig:comp1}, where OIDDR-Net's output successfully mimics the illumination settings of the ground-truth without artifacts.\\
\textbf{Any-to-any Relighting}: We modify our AMIDR-Net by changing the number of input channels and removing the skip connections corresponding to the depth maps and train it on VIDIT 2020 dataset.\ We compare our modified AMIDR-Net with state of the art in Table \ref{tb:comp2}, where SA-AE \cite{10} is the winner of AIM 2020 any-to-any relighting track and \cite{56} is an encoder-decoder network proposed by another participant of the same challenge.\ We also compare our method with an adapted version of \cite{15}, which is originally proposed for portrait relighting. According to Table \ref{tb:comp2}, AMIDR-Net outperforms others w.r.t. all evaluation metrics. Fig. \ref{fig:comp2} visualizes three outputs from different methods where we see how our AMIDR-Net changes the illumination of the input according to guide image without artifacts.\\
\textbf{Comparison with NTIRE 2021 Relighting Methods}: Additionally, we compare our models with two methods from the top 5 methods of the NTIRE 2021 relighting challenge.\ As Table \ref{tb:NTIRE2021} confirms,  OIDDR-Net and AMIDR-Net are among the top-performing methods. OIDDR-Net
ranked second in one-to-one relighting in terms of MPS
and AMIDR-Net ranked second in terms of PSNR.\smallsqueezeup 
\begin{table}[]
\caption{Comparison with state of the arts for any-to-any relighting on VIDIT'20 validation set. (LPIPS and MPS are not made available by other works. The runtime for \cite{56} is not reported.)} 
\label{tb:comp2}\centering
\resizebox{.4\textwidth}{!}{
    \begin{tabular}{c|c|c|c}
    \hline
         Model&PSNR&SSIM&Runtime(s)  \\
         \hline
         AMIDR-Net (ours)& \textbf{19.16}&\textbf{0.6621}&0.51\\
         \hline
         SA-AE \cite{10}&18.06&0.6480&0.15\\
         \hline
         DPR \cite{15}&16.40&0.5238&0.095\\
         \hline
         Dong \emph{et al.} \cite{56}&18.07&0.5994&---\\
         \hline
    \end{tabular}}
\end{table}
\begin{table}[]
\caption{Comparison with other methods in NTIRE2021 relighting challenge on VIDIT'21 test set.}
\label{tb:NTIRE2021}\centering
\resizebox{.5\textwidth}{!}{
    \begin{tabular}{c|c|c|c|c|c|c}
    \hline
         Track&Model&PSNR&SSIM&LPIPS&MPS&Runtime(s)  \\
         \hline
         \multirow{3}{*}{One-to-one}&OIDDR-Net (ours)& 18.83&0.6874&0.1634&0.7620&\textbf{0.53}\\
         \cline{2-7}        & Method 1&19.14&0.6931&0.1605&0.7663&2.88\\
         \cline{2-7}
         &Method 2&18.27&0.6772&0.1670&0.7551&2.12\\
         \hline
         \multirow{3}{*}{Any-to-any}&AMIDR-Net (ours)&\textbf{20.14}&0.6711&0.2028&0.7341&\textbf{0.51}\\
         \cline{2-7}
         &Method 1&19.22&0.6784&0.1566&0.7609&2.04\\
         \cline{2-7}
         &Method 2&18.60&0.6508&0.1661&0.7423&0.6740\\
         \hline
    \end{tabular}}
    \squeezeup
\end{table}
\section{Conclusion}\smallsqueezeup 
\noindent We develop a physically inspired dense fusion network
for image relighting. Our method benefits from the capability
of dense networks in extracting representative features, while simultaneously estimating albedo and shading
– key components of the relighting physical model.
The simultaneous intrinsic image decomposition and direct
relighting help the model refine its feature extraction
by joint optimization. This leads to physically more feasible
results in terms of illumination parameters and therefore
less artifacts in the obtained relighted images. Ablation
studies explain the role of each component in models
proposed both for one-to-one and any-to-any relighting.
Comparisons with existing literature on benchmark datasets
and competing methods in the NTIRE’21 relighting challenge
show our proposal achieves state-of-the-art results.

{\small
\bibliographystyle{ieee_fullname}
\bibliography{cvpr}
}

\end{document}